\documentclass{article}

\usepackage{microtype}
\usepackage{graphicx}
\usepackage{subfigure}
\usepackage{booktabs} 

\usepackage{hyperref}

 \usepackage[accepted]{icml2023}

\usepackage{amsmath}
\usepackage{amssymb}
\usepackage{mathtools}
\usepackage{amsthm}

\usepackage[capitalize,noabbrev]{cleveref}

\theoremstyle{plain}
\newtheorem{theorem}{Theorem}[section]

\theoremstyle{definition}
\newtheorem{definition}[theorem]{Definition}

\theoremstyle{remark}

\usepackage[textsize=tiny]{todonotes}
\usepackage{natbib}
\usepackage{textcomp} 
\usepackage{siunitx}
\usepackage{adjustbox}
\usepackage{pgfplots}
\usepackage{pgf}
\DeclareMathOperator*{\argmin}{arg\,min}
\newcommand\norm[1]{\left\lVert#1\right\rVert}

\usepackage{ifthen}
\newboolean{comment}
\setboolean{comment}{True} 

\icmltitlerunning{On a Connection between Differential Games, Optimal Control, and Energy-based Models for Multi-Agent Interactions}

\begin{document}

\twocolumn[
\icmltitle{On a Connection between Differential Games, Optimal Control, and Energy-based Models for Multi-Agent Interactions}



\icmlsetsymbol{equal}{*}

\begin{icmlauthorlist}
\icmlauthor{Christopher Diehl}{tu}
\icmlauthor{Tobias Klosek}{tu}
\icmlauthor{Martin Krüger}{tu}
\icmlauthor{Nils Murzyn}{comp}
\icmlauthor{Torsten Bertram}{tu}
\end{icmlauthorlist}

\icmlaffiliation{tu}{Institute of Control Theory and Systems Engineering, TU Dortmund University,  Dortmund, Germany}
\icmlaffiliation{comp}{ZF Friedrichshafen AG, Artificial Intelligence Lab, Saarbrücken, Germany}

\icmlcorrespondingauthor{Christopher Diehl}{christopher.diehl@tu-dortmund.de}

\icmlkeywords{Machine Learning, ICML}

\vskip 0.3in
]



\printAffiliationsAndNotice{}  

\begin{abstract}
Game theory offers an interpretable mathematical framework for modeling multi-agent interactions. However, its applicability in real-world robotics applications is hindered by several challenges, such as unknown agents' preferences and goals. To address these challenges, we show a connection between differential games, optimal control, and energy-based models and demonstrate how existing approaches can be unified under our proposed \textit{Energy-based Potential Game}  formulation. Building upon this formulation, this work introduces a new end-to-end learning application that combines neural networks for game-parameter inference with a differentiable game-theoretic optimization layer, acting as an inductive bias. The experiments using simulated mobile robot pedestrian interactions and real-world automated driving data provide empirical evidence that the game-theoretic layer improves the predictive performance of various neural network backbones. 
\end{abstract}

\section{Introduction} 
Modeling multi-agent interactions is essential for many robotics applications like motion forecasting and control.
For instance, a mobile robot or a self-driving vehicle has to interact with other pedestrians or human-driven vehicles to navigate safely toward its goal locations. Although data-driven approaches have made significant progress in multi-agent forecasting, challenges arise due to the additional verification requirements in safety-critical domains. Hence, \citet{Geiger2021} postulate the following critical objectives among others: (i) Integrating well-established principles like prior knowledge about multi-agent interactions to facilitate effective generalization, (ii) Ensuring interpretability of latent variables in models, enabling verification beyond mere testing of the final output.

Game-theoretic approaches, utilizing differential/dynamic games \cite{DynamicNonCoopGT}, incorporate priors based on physics and rationality, such as system dynamics and agent preferences, into interaction modeling.  Here, non-cooperative game-theoretic equilibria describe interactions, and solvers typically search for local equilibria \cite{AlGames2022, liu2023learning} based on the current observation $\mathbf{o}_0$, resulting in a single (uni-modal) joint strategy $\mathbf{u}$.
While finding a suitable cost parametrization is non-trivial \cite{ Knox2023, Diehl2023} for the robot (e.g., an self-driving vehicle (SDV) in the open world), knowing the preferences and goals of all other agents is an unrealistic assumption. For example, the intents of human drivers are not directly observable. That makes online inference of game parameters, such as goals and cost weights, necessary \cite{Peters2021}.
\begin{figure}
	\includegraphics[width=\columnwidth]{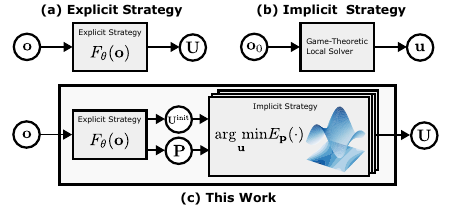}
	\vspace{-0.7cm}
	\caption{Strategy Representations. (a) Learned explicit strategy producing a multi-modal solution $\mathbf{U}$ based on observation $\mathbf{o}$. (b) Implicit game-theoretic strategy, which only considers the current observation $\mathbf{o}_0$ and hence does not account for the prior interaction evolution. Moreover, it has fixed game parameters and produces a local (uni-modal) solution $\mathbf{u}$. (c) This work infers multi-modal strategy initializations $\mathbf{U}^{\textrm{init}}$ and game parameters $\mathbf{P}=\{\mathbf{p}^1,\dots,\mathbf{p}^M\}$ with an explicit strategy, then performs $M$ game-theoretic energy minimizations in \textit{parallel} in a learnable end-to-end framework.}
	\label{FigureOverview}
	\vspace{-0.5cm}
\end{figure}

On the other hand, neural network-based approaches achieve state-of-the-art (SOTA) performance on motion forecasting benchmarks. Works like \cite{TrajectronPP} and \cite{MultiPathPlusPlus} employ \textit{explicit strategies} by utilizing feed-forward neural networks with parameters $\mathbf{\theta}$ to generate multi-modal joint strategies $\mathbf{U} = F_\theta(\mathbf{o})$ based on the observed context $\mathbf{o}$. Although these models achieve impressive results, they are considered as low interpretable black-box models with limited controllability \cite{explain2022}.
\textit{How can we leverage the benefits of both groups of approaches?}
	
Energy-based neural networks \cite{lecun2006tutorial}, a type of generative models, provide an $\textit{implicit}$ mapping
\begin{equation}
		\mathbf{u}^*=\argmin _{\mathbf{u}} E_\theta(\mathbf{u}, \mathbf{o}) .
\end{equation}
\citet{Florence} demonstrate the advantageous properties of such implicit models in single-agent control experiments. This work shows how to parameterize the energy $E_{\theta}(\cdot)$ with a potential game formulation \cite{MONDERER1996124}. Hence, we combine \textit{explicit  strategies} for initialization and parameter inference with \textit{implicit strategies} as shown in Fig. \ref{FigureOverview}. 
\begin{figure}
	\centering
	\begin{tikzpicture}
		\node [anchor=center, inner sep=0] (image) at (0,0) {\includegraphics{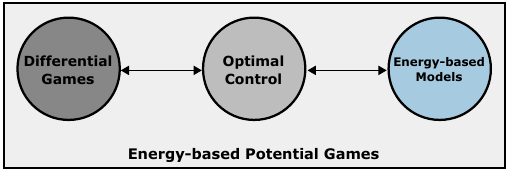}};
		\node [align=center, font=\tiny, text=white, text width=2cm] at (-1.6,-0.4)
		 {\unskip\citeauthor{PotentialDifferentialGames} \citeyear{PotentialDifferentialGames}};
		 \node [align=center, font=\tiny, text=white, text width=2cm] at (+1.6,-0.4)
		 {\cite{EBMIOC}};
	\end{tikzpicture}
	\vspace{-0.5cm}
	\caption{Energy-based Potential Games as a connection of different research areas.}
	\label{Energybasedpotentialgames}
	\vspace{-0.5cm}
\end{figure}

\textbf{Contribution.} This paper contributes the following:
Theoretically, this work proposes \textit{Energy-based Potential Game} (EPO) as a class of methods connecting differential games, optimal control, and energy-based models (EBMs), as visualized in Fig. \ref{Energybasedpotentialgames}. We further show how existing approaches can be unified under this framework. Application-wise, this work proposes a differentiable Energy-based Potential Game Layer (EPOL), which is combined with hierarchical neural network backbones in a novel system architecture.
Third, we demonstrate that our practical implementation improves the performance of different SOTA neural network backbone architectures in simulated and real-world motion forecasting experiments.

\section{Related Work}
\label{relatedwork}

\textbf{Game-Theoretic Planning.}
Game-theoretic motion planning approaches \cite{ILQG2020, AlGames2022, liu2023learning}, aiming to find Nash equilibria  (NE), capture the interdependence about how one agent's action influences other agents' futures. However, these approaches typically involve computationally intensive coupled optimal control problems. Hence, \citet{Geiger2021} and \citet{PotentialILQR} formulate the problem as potential game \cite{MONDERER1996124,PotentialDifferentialGames}, enabling the solution of only a single
 optimal control problem (OCP). 
Different works use filtering techniques \cite{LeCleac} or inverse game-solvers \cite{Peters2021} to learn game parameters. However, these methods have been primarily evaluated in simulation, and the learning objectives of \citet{Peters2021} assumed unimodal distributions.
In contrast, our work utilizes SOTA neural networks to infer cost parameters and strategy initializations end-to-end. 
 We further, provide empirical evidence using a interactive real-world driving dataset.

\textbf{Data-driven Motion Forecasting.} Motion forecasting approaches using neural networks currently represent the SOTA in benchmarks for various applications, such as human vehicle \cite{Ettinger_2021_ICCV} or pedestrian prediction \cite{trajnet}. Most works focus on modeling interactions in the observation encoding part.  For instance, the attention \cite{VN_2020_CVPR} or convolutional social pooling mechanism \cite{convSocialPooling} are commonly employed. \citet{Dsdnet} and \citet{JFP2023} also utilize EBMs, with sampling instead of gradient-based optimization, like in this work. 
 App. \ref{app:related_work} provides a extension of the related motion forecasting applications. In general, our approach is complementary to the developments in motion forecasting, as it can be integrated on top of various network architectures as later shown in Section (\ref{Experiments}).

\textbf{Differentiable Optimization for Machine Learning.}
Advances in differentiable optimization \cite{Amos2017, Theseus} enable our work allowing the combination of optimization problems with learning-based models such as neural networks. 
 \citet{Geiger2021} use a concave maximization-based motion forecasting application, which is restrictive for general real-world scenarios. Additionally, the experiments are limited by the dataset size (max. 25 samples) and only involve two agents. The concurrent work of \citet{liu2023learning} proposes a combination with a differentiable optimization planner evaluated on a simulated dataset and does not account for multi-modal demonstrations and predictions. Both works serve as proof of concepts and utilize simple network architectures with only two hidden layers. By contrast, we account for multi-modal behavior, evaluate on larger real-world datasets, and show that our game-theoretic layer can be easily applied to different SOTA neural networks. Moreover, both approaches draw no connection between EBMs and game theory.

\textbf{Energy-based Model.}
The works of \citet{lecun2006tutorial} and \citet{song2021train} provide reviews for EBMs.
\citet{Belanger2017} identify three main paradigms for energy learning: (i) Conditional Density Estimation, (ii) Exact Energy Minimization (iii) Unrolled Optimization. Models of the first group (i) use the probabilistic interpretation that low-energy regions have high probability. Approaches utilize maximum likelihood estimation (MLE) \cite{song2021train, XieICML2016}, noise contrastive (NC) divergence \cite{Hinton_02}, or NC estimation \cite{NCE} learning objectives.
Type (ii) methods solve the energy optimization problem exactly and differentiate by employing techniques such as the \textit{implicit function theorem} \cite{Amos2017}. Methods from type (iii), like \cite{Belanger2017}, approximate the solution with a finite number of gradient steps and backpropagate through the unrolled optimization.

One closely related application is EBIOC \cite{EBMIOC}, which proposes to use EBMs for inverse optimal control with a type (i) MLE learning. The application of this paper investigates type (iii) methods. Unlike EBIOC, we use SOTA neural network structures and provide deeper analysis in multi-agent scenarios and multi-modal solutions. 
Lastly, EBIOC draws no connections to game theory. However, EBIOC can also be viewed under the EPO formulation (Section \ref{seq:discussionrelatedapplications}) under specific assumptions.

To the author's best knowledge, besides the new connection of the three fields, this work's application is the first to combine nonlinear differentiable game-theoretic optimization with neural networks and successfully demonstrate its performance on considerably large real-world datasets.

\section{Energy-based Potential Games}
\label{sec::Energybasedpotentialgames}
This section describes the EPO framework. After introducing the game-theoretic background based on the works of \citet{DynamicNonCoopGT}, \citet{PotentialDifferentialGames} and \citet{PotentialILQR} in Section \ref{Background}, we will show how to connect the potential game with EBMs in Section \ref{sec::connectiong_pdg_ebm} and discuss how different approaches can be unified under the EPO framework in Section \ref{DiscussionOfRelatedApproache}.
\subsection{Background}
\label{Background}

\textbf{Differential Games.}
Assume we have $N$ agents and $\mathbf{u}_i(t) \in \mathbb{R}^{n_{u,i}}$
 represents the \textit{control} vector and $\mathbf{x}_i(t) \in \mathbb{R}^{n_{x,i}}$ the \textit{state} vector for each agent $i, 1 \leq i \leq N$ at timestep $t$. $n_{u,i}$ and $n_{x,i}$ denote the dimension of the control and state of agent $i$. The overall state evolves according to a time-continuous differential equation with dynamics $f(\cdot)$:
\begin{equation}
	\dot{\mathbf{x}}(t)=f(\mathbf{x}(t), \mathbf{u}(t), t).
	\label{equ:joint_agentdynamics}
\end{equation} starting at the initial state $\mathbf{x}(0) = \mathbf{x}_{0}$.
$\mathbf{u}(t)=\left(\mathbf{u}_1(t), \cdots, \mathbf{u}_N(t)\right)\in \mathbb{R}^{n_{u}}$ and $\mathbf{x}(t)=\left(\mathbf{x}_1(t), \cdots, \mathbf{x}_N(t)\right)\in \mathbb{R}^{n_{x}}$ are the concatenated vectors of all agents controls and states at time $t$, with dimensions $n_u=\sum_{i}n_{u,i}$ and $n_x=\sum_{i}n_{x,i}$. Assume each agent minimizes cost
\begin{equation}
	C_i\left(\mathbf{x}_0, \mathbf{u}\right)=\int_0^T L_i(\mathbf{x}(t), \mathbf{u}(t), t) \mathrm{d}t+S_i(\mathbf{x}(T))
	\label{equ:single_cost_function}
\end{equation} with time horizon $T$, running cost $L_i$ and terminal cost $S_i$ of agent $i$. Costs are assumed to be conflicting rendering the game \textit{noncooperative}.
For instance, in robotics applications, the cost function $C_i$ can be designed to encompass the agents' objectives of reaching a specified goal (encoded as $S_i$), while simultaneously considering collision avoidance and minimizing control efforts (represented by $L_i$).
$\mathbf{u}_i: [0,T] \times \mathbb{R}^{n_{x,i}} \rightarrow \mathbb{R}^{n_{u,i}}$ 
defines an \textit{open-loop strategy}\footnote{Open-loop strategies provide equivalence between strategy and control actions for all time instants \cite{DynamicNonCoopGT}. Hence, for clarity, we omitted to introduce a new variable for the strategy, and overloaded the notation for $\mathbf{u}_i$ such that it describes the controls of agent $i$ in the time interval $[0,T]$.}
and $\mathbf{u}_{-i}$
 defines the open-loop strategy for all players \textit{except} $i$. Then, $\mathbf{u} $
  defines a \textit{joint strategy} for all agents. Let $\mathbf{x}_0$ be the initial measured state. We can now characterize the differential game with notation: $\Gamma_{\mathbf{x}_0}^T:=\left(T,\left\{\mathbf{u}_i\right\}_{i=1}^N,\left\{C_i\right\}_{i=1}^N,f\right)$.
  
Then, let us recall the following definition for NE from  \citet[Chaper~6]{DynamicNonCoopGT}:
\begin{definition}
	\label{def:Nash}
	Given a differential game defined by all agents dynamics (\ref{equ:joint_agentdynamics}), and costs (\ref{equ:single_cost_function}), a joint strategy $\mathbf{u}^*=\left(\mathbf{u}_1^*, \ldots, \mathbf{u}_n^*\right) $
	is called an open-loop Nash equilibrium (OLNE) if, for every $i=1, \ldots, N$,
\end{definition} \vspace{-0.5cm} \begin{equation}
C_i\left(\mathbf{u}^*\right) \leq  	C_i\left(\mathbf{u}_i, \mathbf{u}_{-i}^*\right)\,\,  \forall \, \mathbf{u}_i, 
\end{equation} where $(\mathbf{u}_i, \mathbf{u}^*_{-i})$ is a shorthand for $\left(\mathbf{u}_1^*, \ldots, \mathbf{u}_{i-1}^*, \mathbf{u}_i, \mathbf{u}_{i+1}^*, \ldots, \mathbf{u}_N^*\right)$. Intuitively speaking, no agent is incentivized to unilaterally change its strategy, assuming that all other agents keep their strategy unchanged. 

\textbf{Potential Differential Games.}
Finding a NE involves solving \textit{N-coupled} OCPs, which is non-trivial and computationally demanding \cite{Geiger2021, PotentialILQR}. However, according to \citet{PotentialDifferentialGames} there exists a class of games, namely \textit{potential differential games} (PDGs), in which only the solution of a \textit{single} OCP is required, and its solutions correspond to OLNE of the original game. 
\begin{definition}
	(cf. \citet{PotentialDifferentialGames}) A differential game $\Gamma_{x_0}^T$,  is called an open-loop PDG if there exists an OCP such that an open-loop optimal solution of this OCP is an OLNE for $\Gamma_{x_0}^T$.
\end{definition}
Theorem 1 from \citet{PotentialILQR} (see also App. \ref{app:theorem}) implies, under the assumption of decoupled dynamics \begin{equation}
	\dot{\mathbf{x}}_i(t)=f(\mathbf{x}_i(t), \mathbf{u}_i(t), t) \quad \forall \,i,
	\label{equ:single_agentdynamics}
\end{equation}
 that such an OCP\footnote{The assumption of decoupled dynamics seems reasonable in interactive (robot) trajectory planning settings, as the coupling between agents mainly occurs due to the coupling of agents' cost functions, such as collision avoidance \citet{PotentialILQR}.} is given by:
\begin{align}
	\label{equ:PotentialOCP}
	&\min _{\mathbf{u}(\cdot)} \int_0^T p(\mathbf{x}(t), \mathbf{u}(t), t) \mathrm{d} t+\bar{s}(\mathbf{x}(T))   \\
	&\text{subject to Equation } (\ref{equ:single_agentdynamics}). \nonumber
\end{align}
Here, $p(\cdot)$ and $\bar{s}(\cdot)$ are so called \textit{potential functions}. It is further shown that in the context of interactive game-theoretic trajectory planning the potential function cost terms of the agents have to be composed of two terms: (i) Cost terms $C_i^{\textrm{own}}(\mathbf{x}_i(t), \mathbf{u}_i(t) )$ that only depend on the state and control of agent $i$  (e.g., tracking costs or control input costs) and (ii) pair-wise coupling terms $C_{i,j}^{\textrm{pair}}(\mathbf{x}_i(t), \mathbf{x}_j(t))$ between agents $i$ and $j$, which could encode some common social norms, such as collision avoidance. Further, the coupling terms have to fulfill the property (Theorem 2 \citet{PotentialILQR}): $C_{i,j}^{\textrm{pair}}(\mathbf{x}_i(t), \mathbf{x}_j(t)) = C_{j,i}^{\textrm{pair}}(\mathbf{x}_i(t), \mathbf{x}_j(t)) \, \forall \,  i \neq j$. Intuitively speaking, two agents $i$ and $j$ care the same for common social norms. The potential functions are then given by 
	\begin{align}
		\nonumber p(\cdot) & = \sum_{i=1}^{N} C_i^{\textrm{own}}(\mathbf{x}_i(t), \mathbf{u}_i(t) )+ \sum^N_{1\leq i<j} C_{i,j}^{\textrm{pair}}(\mathbf{x}_i(t), \mathbf{x}_j(t))    \\
		\bar{s}(\cdot) & = \sum_{i=1}^{N} C_{i,T}^{\textrm{own}}(\mathbf{x}_{i}(T) ).
		\label{equ:multi_agent_OCP_cost}
	\end{align}

\subsection{Connecting Potential Differential Games with Energy-based Models}
\label{sec::connectiong_pdg_ebm}
While PDGs provide more tractable solutions to the original game, challenges still arise due to unknown game parameters, like preferences for tracking costs or common social norms. Hence, this work aims to infer the parameters online using function approximators, such as neural networks, based on an observed context $\textbf{o}$ (e.g., agents' histories, map information, or raw-sensor data in robotics applications). We now demonstrate how to connect PDGs to EBMs, laying the foundation for unifying various existing applications (Section \ref{seq:discussionrelatedapplications}) and our practical solution in Section \ref{PracticalImplementation}. 

\textbf{Direct Transcription.} 
Due to its simplicity and resulting low number of optimization variables, we apply single-shooting, a direct transcription method \cite{2001_John}, to transform the time-continuous formulation of (\ref{equ:single_agentdynamics}) and (\ref{equ:PotentialOCP}) into a discrete-time OCP. Let the discretized time interval be $\left[0,T \right]$ with $0=t_0 \leq t_1 \leq \dots \leq t_k \leq \dots \leq t_K = T$ and $k=0,\dots, K$. We assume a piecewise constant control $\textbf{u}_i(t_k):=\textbf{u}_{i,k}=\text{constant}$ for $ t\in\left[t_k,t_k+\Delta t \right)$ , where $\Delta t=t_{k+1}-t_{k}$ denotes the time interval. Assume an approximation of the system dynamics (\ref{equ:single_agentdynamics}) by an explicit integration scheme with $\mathbf{x}_{i,k+1} = f(\mathbf{x}_{i,k}, \mathbf{u}_{i,k})$. By applying single-shooting, the state $\mathbf{x}_i(t_k):=\mathbf{x}_{i,k}$ is obtained by integrating the system dynamics based on the controls $\mathbf{u}_{i,k}$ for $k=0,\dots, K-1$. Hence, states $\mathbf{x}_{i,k}$ of agent $i$ are a function of the initial (measured) agent state $\mathbf{x}_{i,0}$ and the strategy $\mathbf{u}_i \in \mathbb{R}^{n_{u,i} \times K}$.

\textbf{EPO Optimization Problem.}
Let us now formulate the solution of the resulting discrete-time OCP given by: 
\begin{equation}
	\mathbf{u}^*=\argmin _{\mathbf{u}} \sum_{k=0}^{K-1} p(\mathbf{x}_k, \mathbf{u}_k) + \bar{s}(\mathbf{x}_K),
	\label{equ:discrete_time_ocp_cost}
\end{equation} with discrete-time joint state $\mathbf{x}_k \in \mathbb{R}^{n_x}$ and control $\mathbf{u}_k \in \mathbb{R}^{n_u}$ at timestep $k$ and joint strategy $\mathbf{u}\in \mathbb{R}^{n_u \times K}$.
Now assume inference of the game parameters $\mathbf{p}=\phi_\theta(\mathbf{o})$  based on observations $\mathbf{o}$, and we can interpret the cost as an energy function, similar to \citet{EBMIOC}. Hence, the cost terms are now functions of the observations and also depend on some learnable parameters $\theta$. That leads to the energy optimization problem
\begin{equation}
	 \mathbf{u}^*=\argmin_{\mathbf{u}} \sum_{i=1}^{N} E^{\textrm{own}}_{\mathbf{p},i}(\mathbf{u}_i, \mathbf{o}) + \sum^N_{1\leq i<j} E_{\mathbf{p},i,j}^{\textrm{pair}}(\mathbf{u}_i, \mathbf{u}_j, \mathbf{o}).
	 \label{equ:energy_opt_problem}
\end{equation}  Here we combined Equations (\ref{equ:multi_agent_OCP_cost}) and (\ref{equ:discrete_time_ocp_cost}). 
Remember that states are functions of the strategy (sequence of controls) and the initial observation. Hence, state arguments are omitted. $E^{\textrm{own}}_{\mathbf{p},i}(\mathbf{u_i}, \mathbf{o})$ represent an agent specific energy, which can contain running and terminal costs,  and $E_{\mathbf{p},i,j}^{\textrm{pair}}(\mathbf{u}_i, \mathbf{u}_j, \mathbf{o})$ an pairwise interaction energy, whereas both are summed over all $K$ timesteps. The energies could depend on $\textbf{o}$ in two ways, explicitly and implicitly, through the inferred parameters $\textbf{p}$\footnote{The energies of our implementation (Section \ref{PracticalImplementation}) depend on $\textbf{o}$ through $\mathbf{p}=\phi_\theta(\mathbf{o})$. Moreover, energy features dependent on the states are functions of the initial state $\mathbf{x}_0$ extracted from $\mathbf{o}_0$, which induces another dependence on the observations.}. The interpretation as an energy, now allows to apply EBMs techniques for learning.

\subsection{Discussion of Related Applications}
\label{seq:discussionrelatedapplications}
This section revisits the literature and shows how existing applications from the field of multi-agent forecasting can be viewed under the EPO framework. Table \ref{tab:comparison} provides a comparison in terms of the energy structure, the method for solving the energy optimization problem (\ref{equ:energy_opt_problem}), and the learning type (see Section \ref{relatedwork} EBMs). These works provide additional empirical evidence that modeling real-world multi-agent interactions as PDG is promising. \textit{Note that none of these approaches draw connections between PDGs, optimal control, and EBMs.}
\begin{table}[!t]
	\vspace{0.0cm}
	\centering
	\caption{A comparison of EPO applications. MB: Model-based. GB: Gradient-based. CDE: Conditional Density Estimation. IFT: Implicit Function Theorem. UN: Unrolling. *Assuming the fully-connected graph of  \citet{JFP2023}. $\diamond$ TGL uses hard inequality constraints to force a forward movement of the agents, whereas the proposed EPO framework penalizes inequalities in the energy functions. $\star$ Assuming that the cost follows  the PDG formulation.}
	\resizebox{\columnwidth}{!}{
		\begin{tabular}{l c c c c c c c c c c c c}
			\toprule

			&Energy & Energy & Learning     \\
			&Structure & Optimization & Type    \\
			\midrule
			\citet{JFP2023}* & NN & Sampling (Learn.) &  CDE   \\
			\citet{Dsdnet} & Nonlin.+NN & Sampling (MB)  &  CDE   \\
			\midrule
			\citet{Geiger2021}$\diamond$  & Convex & GB & IFT    \\
			\citet{EBMIOC}$\star$ & Nonlin./NN & GB & CDE\\
			\midrule
			Section (\ref{PracticalImplementation}) & Nonlin. & GB &  UN 
			\\
			\bottomrule	
	\end{tabular}}
	\label{tab:comparison}
	\vspace{-0.5cm}
\end{table}

DSDNet \cite{Dsdnet} and JFP \cite{JFP2023} use neural networks to approximate the energy and optimize via sampled future states. These future states can be generated by unrolling the dynamics with controls generated by model-based or learning-based sampling. 
 Both works use the probabilistic interpretation of EBMs with a conditional density given by $p_\theta(\mathbf{u} | \mathbf{o}) = \frac{1}{Z}=\exp(-E_\theta(\mathbf{u},\mathbf{o}))$ learned by MLE, with network parameters $\theta$, and normalization constant $Z$, which is often intractable to compute in closed-form and needs to be approximated. 
In contrast, TGL (\citet{Geiger2021}) uses the implicit function theorem to learn the energy, which requires convergence to an optimal solution $\mathbf{u}^*$ \cite{Theseus}. The energy has the structure of a linear combination of features $E_\mathbf{w}(\mathbf{u},\mathbf{o}) = \sum_{j} \mathbf{w}_j c_j(\mathbf{o},\mathbf{u})=\mathbf{w}^\intercal c(\mathbf{o},\mathbf{u})$. $\mathbf{w}$ describes the vector of inferred cost/energy function weights, and $c_j(\cdot)$ the cost/energy features. The approach uses gradient-based convex minimization (concave maximization), whereas the required convexity of $c$ is restrictive for general real-world scenarios (e.g., curvy lanes). In contrast, Section \ref{PracticalImplementation} proposes a non-convex nonlinear gradient-based solution. Further, our approach learns by backpropagation through unrolled nonlinear optimization problems. 
\citet{EBMIOC} does not make a potential game assumption in their multi-agent control experiments, nor do the authors draw connections between EBMs and PDGs. However, we also classified their approach under the EPO framework for completeness. 
Neural networks as approximations of the energy are more expressive and can overcome the design of features, which is necessary for domains like SDV \cite{Naumann2020} to approximate human behavior. On the other hand, linear combinations of features allow to incorporate domain knowledge into the training process and provide a level of interpretability \cite{explain2022,liu2023learning}. 
\label{DiscussionOfRelatedApproache}

\section{Practical Implementation}
\label{PracticalImplementation}
This section introduces a practical implementation that integrates a differentiable EPO formulation into the training process of neural networks for multi-agent forecasting.
\begin{figure*}
	\includegraphics{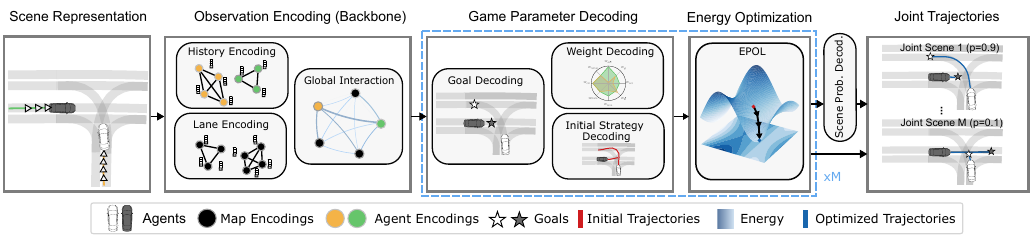}
	\vspace{-0.7cm}
	\caption{System Architecture of the implementation. Agent histories and lanes are encoded using a vectorized representation. The resulting features capture agent-to-agent and agent-to-lane interactions in a global interaction graph. These features are then passed through different decoders. To handle multi-modality in demonstrations, the decoders predict context-dependent goal positions, initial strategies, and weights for energy parameterization. Parallel energy optimization problems are solved in the \textit{inner loop}, resulting in $M$ joint strategies. The dotted blue box represents the parallelization of modes. Unrolling the dynamics generates scene-consistent future joint trajectories. Additionally, a probability decoder predicts scene probabilities. During training, the learnable parameters are updated in the \textit{outer loop} optimization problem based on a multi-task loss.}
	\label{SystemArchitecture}
	\vspace{-0.5cm}
\end{figure*}

\textbf{Problem Formulation.}
We assume access to an object-based representation of the world consisting of agent histories and (optional) map information as visualized for an SDV example in Fig. \ref{SystemArchitecture}. Let an observation $\mathbf{o} = \{\mathbf{h},\mathbf{m}\}$ be defined by a sequence  of \textit{all agents} historic 2-D positions $(x,y)$, denoted by $\mathbf{h}$, with length $H$, and by an optional high-definition map $\mathbf{m}$. Our goal is to predict future multi-modal joint strategies $\mathbf{U}\in \mathbb{R}^{n_u \times K \times M}$ and the associate scene-consistent future joint states $\mathbf{X}\in \mathbb{R}^{n_x \times K \times M}$ of all agents and probabilities $\mathbf{PR} \in [0,1]^{M}$ for all $M$ joint futures. 

\textbf{General Approach.} $\mathbf{U}$ represent $M$ different joint strategies $\mathbf{u}^m \in \mathbb{R}^{n_u  \times K}$ with modes $m = 1,\ldots,M $. $\mathbf{U}$ is obtained by $M$ parallel gradient-based optimizations of energies defined by Equation (Eq.) (\ref{equ:energy_opt_problem}). As minimizing the nonlinear energy with gradient-based solvers can induce problems with local optima, we propose to learn initial strategies $\mathbf{U}^{\textrm{init}} \in \mathbb{R}^{n_u  \times K \times M}$ with a neural network consisting of $M$ strategies $\mathbf{u}^{\textrm{init},m} \in \mathbb{R}^{n_u \times K}$. In addition, for every mode, the network predicts parameter vectors $\mathbf{p}^m \in \mathbb{R}^{n_p}$ with dimension $n_p$, whereas $\mathbf{P}\in \mathbb{R}^{n_p \times M}$ describes the parameters of all modes. Concretely, $\mathbf{p}^m$ contains the weights $\mathbf{w}^m \in \mathbb{R}^{n_w}$ and goals $\mathbf{g}^m \in \mathbb{R}^{2 \times N}$ of all agents. Algorithm \ref{alg:Multi_Agent_Forecasting} provides a pseudocode of the the training and inference procedure.
\begin{minipage}[t]{\columnwidth}
	\vspace{-0.5cm}
	\begin{algorithm}[H]
		\caption{Multi-Agent Forecasting}
		\label{alg:Multi_Agent_Forecasting}
		\begin{algorithmic}[1]

			\REQUIRE observation $\mathbf{o} = \{\mathbf{h},\mathbf{m}\}$, ground truth future joint trajectory $\mathbf{x}_\textrm{GT}$
			\STATE $\mathbf{z}=\phi^{\textrm{glob}}(\phi^{\textrm{agent}}(\mathbf{h}),\phi^{\textrm{lane}}(\mathbf{m}))$ \COMMENT{observation encod.}
			\FOR{$1 \leq i \leq N$} 
			\STATE $\mathbf{G}_i=\phi^\textrm{goal}(\mathbf{z}_i)$  \COMMENT{goal decod.}
			\STATE $\mathbf{W}^{\textrm{own}}_i=\phi^\textrm{own}(\mathbf{z}_i)$  \COMMENT{agent weight decod.}
			\STATE $\mathbf{U}_i^{\textrm{init}}=\phi^\textrm{init}(\mathbf{z}_i)$  \COMMENT{initial strategy decod.}
			\ENDFOR \COMMENT{in parallel for all agents $i$}
			\STATE $\mathbf{W}^{\textrm{pair}}=\phi^\textrm{pair}(\mathbf{z}^{\textrm{all}})$  \COMMENT{interaction weight decod.}
			\FOR{$1 \leq m \leq M$}
			\STATE Parameterize energies (\ref{equ:energy_type_own}), (\ref{equ:energy_type})
			 with $\mathbf{g}^m$, $\mathbf{w}^m$
			\STATE Initialize optimization with $\mathbf{u}^m$  
			\STATE Gradient-based minimization of (\ref{equ:energy_opt_problem})
			\ENDFOR \COMMENT{in parallel for all modes $m$}
			\STATE $\mathbf{X} = f(\mathbf{x}_0,\mathbf{U})$ \COMMENT{unroll dynamics}
			\STATE $\mathbf{PR}=\phi^{\textrm{prob}}(\mathbf{z}^{\textrm{prob}})$ \COMMENT{scene prob. decod.}
			\IF {Training}
			\STATE Update parameters $\theta$ based on $\nabla \mathcal{L}$ (\ref{equ::loss}) 
			\ENDIF
		\end{algorithmic}
	\end{algorithm}
	\vspace{0cm}
\end{minipage}
\subsection{Observation Encoding}
Given observations $\mathbf{o}$, the first step is to encode agent-to-agent and agent-to-lane interactions. Inspired by \citet{VN_2020_CVPR}, this work uses different hierarchical graph neural network \textit{backbones} for observation encoding. We first construct polylines $\mathcal{P}$ based on a vectorized environment representation of the agent histories and map elements.
The resulting subgraphs are encoded with separate \textit{agent history} $\phi^{\textrm{hist}}$ and \textit{lane encoders} $\phi^{\textrm{lane}}$, followed by a network to model high-level interactions in a global graph $\phi^{\textrm{glob}}$. The result is an updated latent polyline feature vector $\mathbf{z}$. Section \ref{Experiments} and App. \ref{app:network_structure} provide additional information.

\subsection{Game Parameter Decoding}
 Let $\mathbf{z}_i$ be the updated feature of agent $i$ after the global interaction graph extracted from $\mathbf{z}$. Next, we will describe the different decoders of the game parameters $\mathbf{P}$ and initial strategies $\mathbf{U}_{\textrm{init}}$ \, which are all implemented by multilayer perceptrons (MLP).
 
\textbf{Goal Decoding.}
Human navigation is partially determined by goals \cite{WOLBERS2010138}. Hence, the \textit{goal decoder} aims to provide a distribution $p^{\textrm{goal}}_i(\mathbf{g}_i|\mathbf{o})$ of future 2-D goal positions $\mathbf{g}_i \in \mathbb{R}^2$. In our SDV experiments, we follow \cite{TNT} and model $p^{\textrm{goal}}_i(\cdot)$ with a categorical distribution over $G$ discrete goal locations to account for multi-modality over agent intents (e.g., lane keeping vs. lane changing). We extract multiple goal positions per agent $\mathbf{G}_i\in\mathbb{R}^{2 \times M}$ by selecting the top $M$ goals from $p^{\textrm{goal}}_i(\cdot)$. $\mathbf{G}_i$ is used to parameterize goal-related features\footnote{Note that the proposed method is not restricted to energies using goal-related features. However, these features can improve the predictive performance, as shown later (Tab. \ref{tab:exid_ablation}).} of energy (\ref{equ:energy_type_own}). 

\textbf{Weight Decoding.}
The energy structure (\ref{equ:energy_type_own}),(\ref{equ:energy_type}) follows a linear combination of features with multi-modal weights $\mathbf{W}\in \mathbb{R}^{n_w \times M}$, which is the concatenation of all agents time-invariant self-dependent weights $\mathbf{W}_i^{\textrm{own}}$ and pairwise weights $\mathbf{W}^{\textrm{pair}}$. These are predicted by two weight decoders. The \textit{agent weight decoder} $\mathbf{W}_i^{\textrm{own}}=\phi^{\textrm{own}}(\mathbf{z}_{i})$ predicts the weights $\mathbf{W}_i^{\textrm{own}}$, based on the agent features $\mathbf{z}_{i}$ from a \textit{single} agent $i$. The \textit{interaction weight decoder} $\mathbf{W}^{\textrm{pair}}=\phi^\textrm{pair}(\mathbf{z}^{\textrm{all}})$ predicts all pairwise weights $\mathbf{W}^{\textrm{pair}}$  at once based on the input $\mathbf{z}^{\textrm{all}}$, which is the concatenation of the agent features $\mathbf{z}_{i}$  from \textit{all} $N$ agents.

\textbf{Initial Strategy Decoding.}
It is important to note that gradient-based methods may not always converge to global or local optima. However, these methods can be highly effective when the solver is initialized close to an optimum. \cite{DC3}. Hence, we learn $M$ initial joint strategies, denoted by $\mathbf{U}^{\textrm{init}}$. More concretely, the \textit{initial strategy decoder} predicts $\mathbf{U}_i^{\textrm{init}}=\phi^\textrm{init}(\mathbf{z}_i)$. The parameters of the goal, agent weight, and strategy decoders are shared for all agents. Hence, the computation is parallelized. 
\subsection{Energy-based Potential Game Layer}
\label{EPOL}
The energy-based potential game layer solves the $M$ optimization problems, defined in Eq. (\ref{equ:energy_opt_problem}) in parallel, using the predicted parameters $\mathbf{P}$ and initializations $\mathbf{U}^\textrm{init}$. 

\textbf{Energy Structure.}
The energies from Eq. (\ref{equ:energy_opt_problem}) have the structure of a linear combination of weighted nonlinear vector-valued functions $c(\cdot)$ and $d(\cdot)$ given by
\begin{align}
	E^{\textrm{own}}_{\mathbf{p},i}(\mathbf{u}^m_i, \mathbf{o}) &= \frac{1}{2} \norm{ (\mathbf{w}^{\textrm{own},m}_i)^\intercal c(\mathbf{u}^m_i,\mathbf{g}^m_i)}^2, \label{equ:energy_type_own}\\
	E_{\mathbf{p},i,j}^{\textrm{pair}}(\mathbf{u}^m_i, \mathbf{u}^m_j, \mathbf{o}) &= \frac{1}{2}\norm{ (\mathbf{w}^{\textrm{pair},m}_{i,j})^\intercal d(\mathbf{u}^m_i, \mathbf{u}^m_j)}^2,
	\label{equ:energy_type}
\end{align}
which allows incorporating domain knowledge into the training process. $\mathbf{w}_{i}^{\textrm{own},m}$, $\mathbf{w}_{i,j}^{\textrm{pair},m}$, and $\mathbf{u}_i^m\in \mathbb{R}^{n_u \times K}$ describe the weight vectors and strategies of mode $m$ and agent $i$. $c(\cdot)$ includes agent-dependent costs, which, for example, can induce goal-reaching behavior while minimizing control efforts. $d(\cdot)$ is a distance measure between two agent geometries. The use of weighted features provides an additional layer of interpretability according to the definition \cite{explain2022}. For instance, a visualization of feature weights provides further insights into the decision-making process. A high weight for reaching a goal lane could indicate a lane change.
Remember from Section \ref{sec::connectiong_pdg_ebm} that the future joint states  $\textbf{x}_i^m$ are a function of strategy $\mathbf{u}_i^m$, connected by the explicit integration scheme of the dynamics. The approach uses an Euler-forward integration scheme with dynamically-extended differentiable unicycle 
dynamics (see App. \ref{app:dynamics}) to model pedestrians, mobile robots, or vehicles.

\textbf{Differentiable Optimization.}
The structure of (\ref{equ:energy_type_own}), (\ref{equ:energy_type}) allows us to solve parallel optimizations using the differentiable Nonlinear Least Square solvers of \citet{Theseus}. The implementation uses the second-order Levenberg–Marquardt method \cite{LeastSquareWright}. Hence, we minimize Eq. (\ref{equ:energy_opt_problem}) by iteratively taking $S$ steps
	$\mathbf{u}_{s+1} = \mathbf{u}_s + \alpha \Delta \mathbf{u}$.
$s$ describes the iteration index with $s=1,\dots, S$ and  $\alpha$ is a stepsize $ 0 < \alpha \leq 1$. $\Delta \mathbf{u}$ is found by linearizing the energy around the current joint strategy $\mathbf{u}$ and subsequently solving a linear system \cite{Theseus}. During training, we can then backpropagate gradients through the unrolled \textit{inner loop energy minimization} based on a loss function of the \textit{outer loop loss minimization}. 

\subsection{Scene Probability Decoding}
The result of the $M$ parallel optimizations are  multi-modal joint strategies $\mathbf{U}$. Unrolling the dynamics leads to $M$ multi-modal future joint state trajectories $\mathbf{X}$ and the goal of the scene probability decoder is to estimate probabilities for each future $\mathbf{PR}=\phi^{\textrm{prob}}(\mathbf{z}^{\textrm{prob}})$. The decoder takes as input the concatenation of $\mathbf{z}^{\textrm{all}}$ and joint trajectories $\mathbf{X}$, denoted by $\mathbf{z}^{\textrm{prob}}$ and outputs probabilities $\mathbf{PR}$ for the $M$ futures.
\subsection{Training Objectives}
\label{TrainingObjectives}

The implementation follows prior work \cite{SceneTransformer, TNT} and minimizes the multi-task loss 
\begin{equation}
	\mathcal{L} = \lambda_1\mathcal{L}^{\textrm{imit}} + \lambda_2\mathcal{L}^{\textrm{goal}} + \lambda_3 \mathcal{L}^{\textrm{prob}}.
	\label{equ::loss}
\end{equation}
 with scaling factors $\lambda_1, \lambda_2, \lambda_3$ of the different loss terms. The imitation loss $\mathcal{L}^{\textrm{imit}}$ is a distance of the joint future closest to the ground truth. As $\mathcal{L}^{\textrm{imit}}$ induces  imitation behavior, the energies/costs will be learned such that solving the optimal control problem with the learned energies/costs results in multi-agent imitation\footnote{We can interpret cost learning as a type of multi-agent inverse reinforcement learning (RL) as \citet{Mehr2023}, sometimes also called multi-agent inverse optimal control \cite{Neumeyer}. The forward pass is a type of multi-agent model-based RL, utilizing planning with learned cost \cite{Moerland2023}.}. $\mathcal{L}^{\textrm{goal}}$ computes the negative log-likelihood based on the predicted goals $\mathbf{G}$ locations and $\mathcal{L}^{\textrm{prob}}$ similarly for the multi-modal future joint states $\textbf{X}$.  
Further details are given in App. \ref{app::losses}. 

\section{Experimental Evaluation}
\label{Experiments}
The experiments presented below aim to answer the following research questions: \textit{Q1} Is the methodology applicable to different motion forecasting backbones, and does it enhance the predictive performance?
\textit{Q2}: What are the most influential hyperparameters?

\textbf{Evaluation Environments.}
The first dataset contains simulated multi-modal mobile robot pedestrian interaction (\textit{RPI}), constructed based on the implementation of \citet{Peters2020}. \textit{exiD} is a real-world dataset of interactive scenarios, captured by drones at different locations of highway \cite{exiD2022}. The datasets contain 60338 (RPI) and 290735 (exiD) samples respectively. Methods are tasked to predict joint futures of $T=\SI{4}{\second}$. Details are given in App. \ref{app:datasets}.

\textbf{Metrics.}
This work follows standard motion forecasting metrics \cite{Rhinehart2019, SceneTransformer, JFP2023}. The \textit{minADE} calculates the $\text{L}_2$ norm of a \textit{single-agent trajectory} out of $M$ predictions with the minimal distance to the groundtruth. The \textit{minFDE} is similar to the minADE but only evaluated at the last timestep.
The \textit{minSADE} and \textit{minSFDE} are the scene-level equivalents to minADE and minFDE, calculating the $\text{L}_2$ norm between \textit{joint trajectories} and joint futures, as \citet{ImplicitLatentVariabModel}. The recent study of \citet{weng2023joint} highlights the importance of these joint metrics. We further calculate the overlap rate \textit{OR} of the most likely-joint prediction, which measures the scene consistency as described by \citet{JFP2023}. When using marginal prediction methods, the joint metrics (minSADE, minSFDE, OR) are computed by first ordering the single agent predictions according to their marginal probabilities and constructing a joint scenario accordingly.

\textbf{Baselines.}
\textit{Constant Velocity} (Const. Vel.) is a kinematic baseline, achieving good results for predicting pedestrians \cite{ConstantVelocity} or highway vehicles \cite{EBMIOC}. The experiments also utilize the following SOTA architectures as baselines and observation encoding backbones. All methods utilize the lane encoders of \citet{VN_2020_CVPR}.
\textit{V-LSTM}: \citet{Ettinger_2021_ICCV} encode agent histories with an LSTM \cite{LSTM} and a single-stage attention mechanism \cite{Attention} directly models the interactions between agents and lanes.
\textit{HiVT-M}: Inspired by \citet{HiVT}, this slightly modified baseline encodes the agent histories with transformers \cite{Attention} and uses a two-stage attention mechanism. First, the map-to-agent interactions are modeled, and subsequently, the agent-to-agent interactions. 
\textit{VIBES}: This unpublished baseline stands for Vectorized Interaction-based Scene Prediction and uses an LSTM for agent encoding and the previously described two-stage attention mechanism. V-LSTM, HiVT-M, and VIBES use a marginal loss formulation by minimizing the minADE for trajectory regression and classification loss similar to \citet{TNT} to estimate probabilities. However, that could lead to inconsistencies in future trajectories as it approximates a \textit{marginal distribution} over future locations \textit{per actor}.
\texttt{Backbone}+\textit{SC}: To make a fair comparison, we introduce additional baselines that minimize a scene-consistent loss, consisting of the minSADE and the same scene probability loss as our approach ($\mathcal{L}=\mathcal{L}^{\textrm{imit}}+\mathcal{L}^{\textrm{prob}}$ from Eq. (\ref{equ::loss})). Moreover, these baselines predict control values like \citet{DKM}. This approach approximates a \textit{joint distribution} over future locations \textit{per scene}. These approaches are closest to our implementation but, in contrast, do not use parameter decoding, nor the EPOL. 
Grid searches were performed to find the optimal hyperparameters for all baselines to ensure a fair comparison. \textit{The hyperparameters for our method are the same across all backbones} (V-LSTM, VIBES, HiVT-M). 

\textbf{Energy Features and Dynamics.} Both experiments use unicycle dynamics. Without loss of generality, agents' geometries are approximated by a circle of radius $r_i$. Hence $d(\cdot)$ in Eq. (\ref{equ:energy_type}) is a Euclidean point-to-point distance, active when the circles overlap. 
In the RPI experiments, specific features in $c(\cdot)$ penalize deviations from goal locations, high controls and control derivations, velocities, as well as violations of state, control, and control derivation bounds. In the exiD experiments, agent-dependent features also penalize high distances to a reference line of the goal, and differences from a reference velocity, but not state or control bounds. 
App. \ref{app:datasets} and \ref{app:implementation_details} provide further details regarding the environments, network architectures, and implementation.

\begin{table} 
	\centering
	\caption{Predictive performance of the different evaluated methods on the RPI test dataset. ADE, FDE, SADE and FDE are computed as the minimum over $M=2$ predictions in $\left[\SI{}{\metre}\right]$. \textbf{Bold} numbers mark the best result and \underline{underlined} numbers the second best of the group of approaches, which uses the same observation encoding backbone. Lower numbers are better.}  
	\begin{tabular}{l c c c c c }
		
		\toprule & \multicolumn{2}{c}{Marginal $\downarrow$} & \multicolumn{3}{c}{Joint $\downarrow$}\\
		\cmidrule(r){2-3}
		\cmidrule(r){4-6}
		Method   & ADE  & FDE & SADE & SFDE & OR \\
		\midrule
		V-LSTM  & $0.12$ & $0.20$  & $0.13$ & $0.23$ & $0.006$ \\ 
		+ SC& $\underline{0.04}$ & $\underline{0.11}$ & $\underline{0.04}$ & $\underline{0.11}$ & $\underline{0.001}$ \\ 
		+ Ours & $\pmb{0.03}$  & $\pmb{0.08}$ & $\pmb{0.03}$ & $\pmb{0.08}$ & $\pmb{0.000}$ \\  %
		
		\bottomrule
	\end{tabular}
\label{tab:RPI}
\vspace{-0.5cm}
\end{table}
\begin{table} 
\centering
\caption{Predictive performance of the different evaluated methods on the exiD test dataset.  
	The metrics and formatting is the same as in Tab. \ref{tab:RPI}, but $M=5$.	
}
\resizebox{\columnwidth}{!}{  
	
	\begin{tabular}{l c c c c c }
		
		\toprule & \multicolumn{2}{c}{Marginal $\downarrow$} & \multicolumn{3}{c}{Joint $\downarrow$}\\
		\cmidrule(r){2-3}
		\cmidrule(r){4-6}
		Method & $\mathrm{ADE}$ & $\mathrm{FDE}$ & $\mathrm{SADE}$ & $\mathrm{SFDE}$ & OR \\
		
		\midrule
		
		V-LSTM & $1.25$ & $3.64$  & $1.98$ & $3.90$ & $0.031$ \\ 
		+ SC&  $\underline{0.82}$ & $\underline{1.95}$ & $\underline{1.07}$ & $\underline{2.63}$ & $\underline{0.010}$ \\ 
		+ Ours & $\pmb{0.80}$  & $\pmb{1.89}$ & $\pmb{0.99}$ & $\pmb{2.37}$ & $\pmb{0.008}$ \\ %
		
		\midrule

		VIBES & $1.35$  & $\pmb{1.68}$  & $1.93$ & $2.93$ & $0.021$ \\ 
		+ SC & $\textbf{0.73}$  & $\underline{1.71}$ & $\underline{1.01}$ & $\underline{2.47}$ & $\underline{0.007}$\\ 
		+ Ours  & $\underline{0.83}$ & $1.99$ & $\pmb{0.99}$ & $\pmb{2.40}$ & $\pmb{0.006}$ \\ 
		\midrule
		
		HiVT-M  & $1.83$  & $2.02$ & $2.39$ & $2.86$ & $0.013$ \\ 
		+ SC  & $\pmb{0.78}$  & $\pmb{1.90}$ & $\underline{1.04}$ & $\underline{2.57}$ & $\underline{0.008}$ \\ 
	
		+ Ours  & $\underline{0.83}$ & $\underline{1.97}$ & $\pmb{1.03}$ & $\pmb{2.48}$ & $\pmb{0.007}$  \\ 
		
		\midrule
		Const. Vel. & $1.16$ & $2.87$ & $1.16$ & $2.87$ & $0.010$\\ 
		
		\bottomrule
	\end{tabular}
}
\label{tab:exidmaineval}
\vspace{-0.3cm}
\end{table}

\subsection{Does the EPOL improve the performance of different observation encoding backbones?}
\label{inductivebias}
\begin{figure}[!t] 
	\includegraphics[width=\columnwidth]{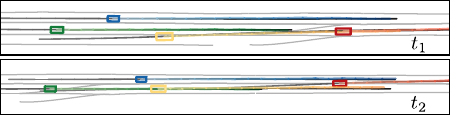}
	\vspace{-0.7cm}
	\caption{Qualitative results in a interactive exiD scenario for two time steps $t_1,t_2$. Agents and the most likely future joint trajectories are visualized in different colors. The saturation increases with the number of predicted steps. The groundtruth (history and future) is visualized with colors from dark grey to black. The map is visualized in light grey and the $x$-axis is about 250 \SI{}{\metre} long.}
	\label{fig::qual_comparison}
	\vspace{-0.2cm}
\end{figure}

Consider the results in Tab. \ref{tab:RPI} and \ref{tab:exidmaineval}, showing the results on both datasets when the approach is applied to different observation encoding backbones. Our implementation consistently outperforms the baselines in all joint distance metrics (minSADE and minSFDE) across all backbones \textit{without backbone-specific hyperparameter tuning}. Note that also the overlap rate decreases due to the game-theoretic inductive bias. Especially joint metrics  are important, as they measure the scene consistency, which is also underlined in a recent study of \citet{weng2023joint}. Further, Fig. \ref{fig::qual_comparison} visualizes qualitative joint predictions in one interactive merging scenarios. Observe how our method produces scene-consistent predictions. For example, the yellow and red car perform a lane change at high speeds. Our model predicts, the resulting interaction accurately. Additional multi-modal predictions are located in App. \ref{app:timeseries}. We conclude that our approach can be applied to different backbones and improves the predictions.

\subsection{Ablation Study}
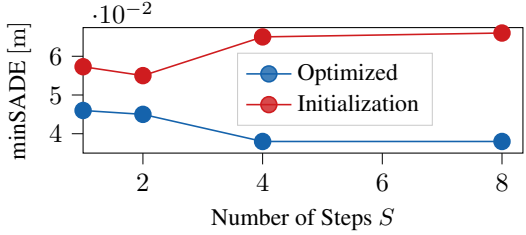
\begin{figure}[!t] 
	\vspace{-0.0cm}
	\centering
\begin{tikzpicture}

\definecolor{darkgray176}{RGB}{176,176,176}
\definecolor{lightgray204}{RGB}{204,204,204}
\definecolor{EPOred}{RGB}{207,28,31}
\definecolor{EPOblue}{RGB}{21,99,172}

\begin{axis}[
legend cell align={left},
legend style={
  fill opacity=0.8,
  draw opacity=1,
  text opacity=1,
  at={(0.35,0.8)},
  anchor=north west,
  draw=lightgray204
},
width=0.9\columnwidth,
height=1.28in,
tick align=outside,
tick pos=left,
x grid style={darkgray176},
xlabel={Number of Steps $S$},
xmin=1, xmax=8.35,
xtick style={color=black},
y grid style={darkgray176},
ylabel={minSADE $\left[\SI{}{\metre}\right]$},
ymin=0.035, ymax=0.0674,
xlabel near ticks,
ylabel near ticks,
label style={font=\footnotesize},
legend style={font=\footnotesize},
ytick style={color=black}
]
\addplot [semithick, EPOblue, mark=*, mark size=3, mark options={solid}]
table {%
1 0.046
2 0.045
4 0.038
8 0.038
};
\addlegendentry{Optimized}
\addplot [semithick, EPOred, mark=*, mark size=3, mark options={solid}]
table {%
1 0.0573
2 0.055
4 0.065
8 0.066
};
\addlegendentry{Initialization}
\end{axis}

\end{tikzpicture}
	\vspace{-0.5cm}
	\caption{Predictive performance of the initial (red) and optimized strategy (blue) as a function of the number of optimization steps on the RPI validation dataset.}
	\label{fig::ablationsteps}	
	\vspace{-0.4cm}
\end{figure}

The experiments identified that the most influential hyperparameter is the number of steps $S$ during optimization. Fig. \ref{fig::ablationsteps} visualizes the dependency. Observe how the approach gets reasonable small metrics with all configurations and hence could be used with different numbers of steps. However, while the distance between the closest optimized joint future and GT gets smaller with increasing optimization steps, the initialization gets slightly pushed away from GT. Hence, with more steps, the approach gets less dependent on the initialization. \citet{huang2023differentiable} observe an similar effect.  

Further ablations for energy features and learned initialization are given in Tab. \ref{tab:exid_ablation}. Turning off the goal-related features inhibits goal-reaching behavior, which is essential for modeling human behavior \cite{WOLBERS2010138,Naumann2020}. Hence, the performance declines in all metrics. When we turn off the learned initialization and initialize the controls for all agents with zeros, we also observe a
decline in performance. However, the decline is lower as unicycle dynamics with controls of zeros correspond to a constant velocity and constant turn rate movement, which is reasonable for highway scenarios. Tab. \ref{tab:exidmaineval} demonstrate that a constant velocity movement is a straightforward yet competitive baseline in highway scenarios, subsequently enhanced through energy optimization. Nevertheless, it fails to attain the performance level exhibited by the learned initialization. The findings in Tab. \ref{tab:exid_ablation} underline the significance of the algorithmic components in this study.

\begin{table} 
	\centering
	\caption{Ablation study investigating the impact of learned initialization, and goal-related features on the exiD test dataset. All models use the V-LSTM backbone encoding. \textbf{Bold} numbers mark the best result and \underline{underlined} numbers the second best of the group of approaches, using the same observation encoding backbone. Lower numbers are better.	
	}
	\begin{tabular}{l c c c c c}
		
		\toprule & \multicolumn{2}{c}{Marginal $\downarrow$} & \multicolumn{3}{c}{Joint $\downarrow$}\\
		\cmidrule(r){2-3}
		\cmidrule(r){4-6}
		Method & $\mathrm{ADE}$ & $\mathrm{FDE}$ & $\mathrm{SADE}$ & $\mathrm{SFDE}$ & $\mathrm{OR}$ \\
		
		\midrule
		\midrule
		Ours (full) & $\pmb{0.83}$ & $\pmb{1.99}$ & $\pmb{0.99}$ & $\pmb{2.40}$ & $\pmb{0.006}$ \\
		No init & $\underline{0.89}$  & $\underline{2.04}$ & $\underline{1.01}$ & $\underline{2.47}$ & $\underline{0.007}$  \\
		No goal & $\underline{0.89}$  & $2.13$ & $1.10$ & $2.64$ & $0.009$	\\

		\bottomrule
	\end{tabular}
	
	\label{tab:exid_ablation}
	\vspace{-0.3cm}
\end{table}

\begin{figure}[!ht] 
	\centering
\begin{tikzpicture}

\definecolor{darkgray176}{RGB}{176,176,176}
\definecolor{darkslategray38}{RGB}{38,38,38}

\begin{axis}[
colorbar,
colorbar style={ylabel={}},
colormap={mymap}{[1pt]
  rgb(0pt)=(0.968627450980392,0.984313725490196,1);
  rgb(1pt)=(0.870588235294118,0.92156862745098,0.968627450980392);
  rgb(2pt)=(0.776470588235294,0.858823529411765,0.937254901960784);
  rgb(3pt)=(0.619607843137255,0.792156862745098,0.882352941176471);
  rgb(4pt)=(0.419607843137255,0.682352941176471,0.83921568627451);
  rgb(5pt)=(0.258823529411765,0.572549019607843,0.776470588235294);
  rgb(6pt)=(0.129411764705882,0.443137254901961,0.709803921568627);
  rgb(7pt)=(0.0313725490196078,0.317647058823529,0.611764705882353);
  rgb(8pt)=(0.0313725490196078,0.188235294117647,0.419607843137255)
},
width=0.9\columnwidth,
height = 1.7in,
point meta max=0.637949511591072,
point meta min=-0.0012355681074162,
x grid style={darkgray176},
label style={font=\footnotesize},
xmin=0, xmax=4,
xtick style={color=black},
xtick={0.5,1.5,2.5,3.5},
xticklabels={2 ,4 ,8 ,12 },
y dir=reverse,
y grid style={darkgray176},
ylabel={Number of Steps $S$},
xlabel={Number of Agents $N$},
ymin=0, ymax=5,
ytick pos=left,
ytick style={color=black},
ytick={0.5,1.5,2.5,3.5,4.5},
yticklabel style={rotate=90.0},
xlabel near ticks,
ylabel near ticks,
yticklabels={1 ,2 ,4 ,8 ,12 }
]
\addplot graphics [includegraphics cmd=\pgfimage,xmin=0, xmax=4, ymin=5, ymax=0] {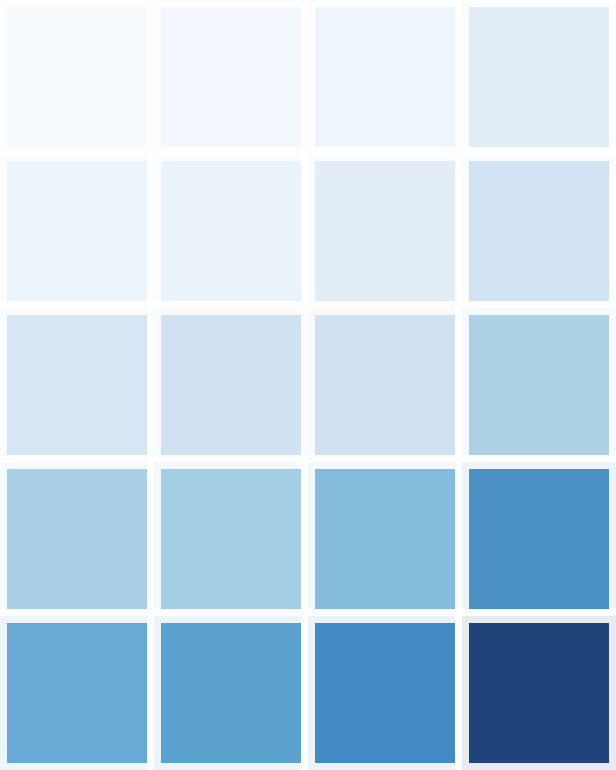};
\draw (axis cs:0.5,0.5) node[
scale=0.7,
text=darkslategray38,
rotate=0.0
]{0.08};
\draw (axis cs:1.5,0.5) node[
scale=0.7,
text=darkslategray38,
rotate=0.0
]{0.09};
\draw (axis cs:2.5,0.5) node[
scale=0.7,
text=darkslategray38,
rotate=0.0
]{0.11};
\draw (axis cs:3.5,0.5) node[
scale=0.7,
text=darkslategray38,
rotate=0.0
]{0.12};
\draw (axis cs:0.5,1.5) node[
scale=0.7,
text=darkslategray38,
rotate=0.0
]{0.10};
\draw (axis cs:1.5,1.5) node[
scale=0.7,
text=darkslategray38,
rotate=0.0
]{0.12};
\draw (axis cs:2.5,1.5) node[
scale=0.7,
text=darkslategray38,
rotate=0.0
]{0.14};
\draw (axis cs:3.5,1.5) node[
scale=0.7,
text=darkslategray38,
rotate=0.0
]{0.14};
\draw (axis cs:0.5,2.5) node[
scale=0.7,
text=darkslategray38,
rotate=0.0
]{0.16};
\draw (axis cs:1.5,2.5) node[
scale=0.7,
text=darkslategray38,
rotate=0.0
]{0.17};
\draw (axis cs:2.5,2.5) node[
scale=0.7,
text=darkslategray38,
rotate=0.0
]{0.21};
\draw (axis cs:3.5,2.5) node[
scale=0.7,
text=darkslategray38,
rotate=0.0
]{0.26};
\draw (axis cs:0.5,3.5) node[
scale=0.7,
text=darkslategray38,
rotate=0.0
]{0.25};
\draw (axis cs:1.5,3.5) node[
scale=0.7,
text=darkslategray38,
rotate=0.0
]{0.27};
\draw (axis cs:2.5,3.5) node[
scale=0.7,
text=darkslategray38,
rotate=0.0
]{0.31};
\draw (axis cs:3.5,3.5) node[
scale=0.7,
text=white,
rotate=0.0
]{0.32};
\draw (axis cs:0.5,4.5) node[
scale=0.7,
text=white,
rotate=0.0
]{0.35};
\draw (axis cs:1.5,4.5) node[
scale=0.7,
text=white,
rotate=0.0
]{0.36};
\draw (axis cs:2.5,4.5) node[
scale=0.7,
text=white,
rotate=0.0
]{0.45};
\draw (axis cs:3.5,4.5) node[
scale=0.7,
text=white,
rotate=0.0
]{0.46};
\end{axis}

\end{tikzpicture}
\begin{tikzpicture}

\definecolor{darkgray176}{RGB}{176,176,176}
\definecolor{darkslategray38}{RGB}{38,38,38}

\begin{axis}[
scaled y ticks=manual:{}{\pgfmathparse{#1}},
width=0.9\columnwidth,
height = 0.8in,
x grid style={darkgray176},
xlabel={Number of Modes $M$},
xmin=0, xmax=5,
xtick pos=left,
xtick style={color=black},
xtick={0.5,1.5,2.5,3.5,4.5},
xticklabels={1,2,4,8,12},
y dir=reverse,
y grid style={darkgray176},
ymajorticks=false,
ymin=0, ymax=1,
ytick style={color=black},
xlabel near ticks,
label style={font=\footnotesize},
ylabel near ticks,
yticklabels={}
]
\addplot graphics [includegraphics cmd=\pgfimage,xmin=0, xmax=5, ymin=1, ymax=0] {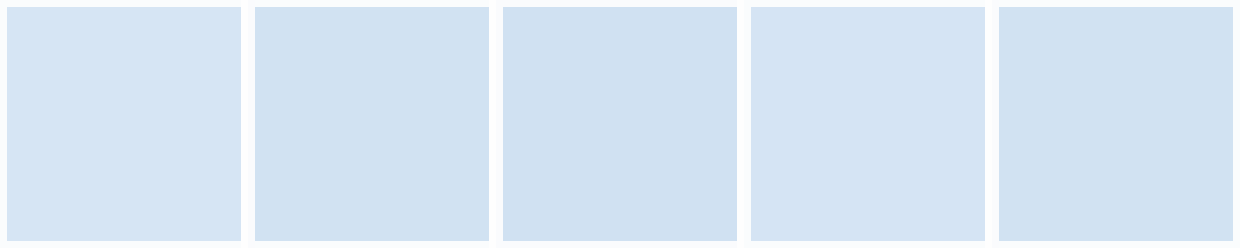};
\draw (axis cs:0.5,0.5) node[
scale=0.6,
text=darkslategray38,
rotate=0.0
]{0.169};
\draw (axis cs:1.5,0.5) node[
scale=0.6,
text=darkslategray38,
rotate=0.0
]{0.170};
\draw (axis cs:2.5,0.5) node[
scale=0.6,
text=darkslategray38,
rotate=0.0
]{0.172};
\draw (axis cs:3.5,0.5) node[
scale=0.6,
text=darkslategray38,
rotate=0.0
]{0.174};
\draw (axis cs:4.5,0.5) node[
scale=0.6,
text=darkslategray38,
rotate=0.0
]{0.175};
\end{axis}

\end{tikzpicture}
	\vspace{-0.2cm}
		\caption{ Mean runtime in [\SI{}{\second}] for different number of agents and optimizations steps averaged over 100 exiD samples with $M=4$. The experiments for the number of modes use $S=2$ and $N=4$.}
	\label{fig::runtime}
	\vspace{-0.5cm}
\end{figure}
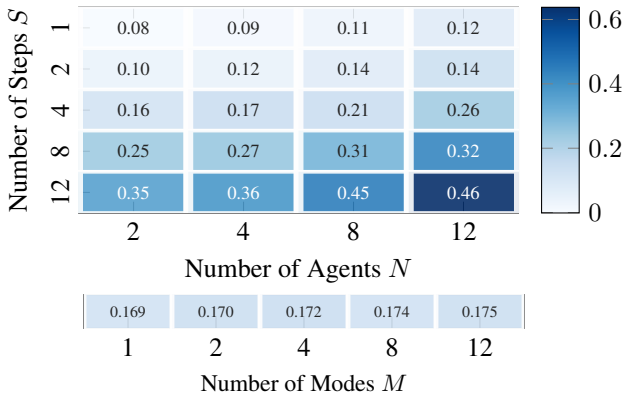

\subsection{Limitations and Future Work.}
As commonly reported in the literature, game-theoretic motion planning approaches suffer from increased runtime, especially with an increasing number of agents. While our implementation scales well with the number of modes (nearly constant runtime), due to parallelization (see Fig. \ref{fig::runtime}), that effect is also present in our non-runtime optimized implementation. However, future work could apply decentralized optimization techniques similar to \citet{williams2023distributed}, to further reduce the runtime. 
 Without loss of generality, our implementation is limited by a fixed number of agents (see App. \ref{app:datasets}) during optimization due to the requirement of fixed-size optimization variables \cite{Theseus}. Future work should overcome this issue and dynamically identify interacting agents, as not all agents constantly interact in a scene. That could be done utilizing the already existing attention mechanism, similar to \citet{Hazard_2022_CVPR}, and would also address the first runtime limitation. Future work should also explore the generalization of our findings to more complex urban settings.
This work considered applications for motion forecasting (open-loop). However, future work could also use our formulation for closed-loop control of one or more agents by executing the most likely trajectory like \citet{Peters2020}.
The proposed EPO framework further open opportunities for various future algorithms, which combine different types of energy structures, optimization, and differentiation techniques as indicated in Section \ref{DiscussionOfRelatedApproache}. 
\section{Conclusions}
This work presented a connection between differential games, optimal control, and EBMs. Based on these findings, we developed a practical implementation that improves the performance of various neural networks in scene-consistent motion forecasting experiments. 
Similiar to \citet{GANIRLEBM}, we hope that by highlighting the connection between these fields, researchers in these three communities will be able to recognize and utilize transferable concepts across domains, particularly in the development of interpretable and scalable algorithms.

\section*{Acknowledgements}
This work was supported by the Federal Ministry for Economic Affairs and Climate Action on the basis of a decision by the German Bundestag and the European Union in the Project \textit{KISSaF - AI-based Situation
Interpretation for Automated Driving.}

\newpage 

\bibliography{example_paper}
\bibliographystyle{icml2023}

\newpage
\appendix
\onecolumn
\section{List of Abbreviations}
\begin{table}[htbp]
	\centering
	\begin{tabular}{ll}
		\textbf{Abbreviation} & \textbf{Description} \\
		\hline
		ADE & average displacement error \\
		CDE & conditional density estimation \\
		Const. Vel.& Constant Velocity \\
		EBM & Energy-based Model \\
		EPOL & Energy-based Potential Game Layer \\
		EPO & Energy-based Potential Game \\
		Eq. & equation \\
		FDE & final displacement error \\
		GB & gradient-based \\
		HiVT-M & Hierarchical Vector Transformer Modified \\
		IFT & implicit function theorem \\
		Learn. &  learned \\
		LSTM & long short-term memory \\
		MG & model-based \\
		MLE & maximum likelihood estimation \\
		NC & noise contrastive \\
		NE & Nash equilibrium \\
		Nonlin. & nonlinear \\
		OLNE & open-loop Nash equilibrium \\
		OR & overlap rate \\
		OCP & optimal control problem \\
		PDG & potential differential game \\
		RPI & robot pedestrian interaction \\
		SADE & scene average displacement error \\
		SDV & self-driving vehicle \\
		SFDE & scene final displacement error \\
		SOTA & state-of-the-art \\
		UN & unrolling \\
		VIBES & Vectorized Interaction-based Scene Prediction \\
		V-LSTM & Vector-LSTM \\
		\hline
	\end{tabular}
\end{table}

\section{Extended Related Work}
\label{app:related_work}
\textbf{Data-driven Motion Forecasting}
Deep learning motion forecasting approaches use different observation inputs. \citet{DKM} uses birds-eye-view images, which have a memory requirement and can lead to discretization errors. \citet{VN_2020_CVPR} propose to use a vectorized environment representation instead, and \citet{Rhinehart2019} uses raw-sensor data. The approaches often utilize an encoder-decoder structure with convolutional neural networks \cite{convSocialPooling}, transformer \cite{SceneTransformer}, or graph neural networks \cite{ImplicitLatentVariabModel} to model multi-agent interactions. In addition to deterministic models \cite{SceneTransformer}, various generative models, such as Generative Adversarial Networks (GANs) \cite{SocialGan} and Conditional Variational Autoencoder formulations  \cite{TrajectronPP}, as well as Diffusion Models \cite{Gu_2022_CVPR}, is used. First, predicting goals in hierarchical approaches like \cite{TNT}, further increases the predictive performance using domain knowledge of the map information. Motion forecasting models can also be conditioned on the control \cite{Diehl2023} or future trajectory \cite{TrajectronPP} of one agent. However, these conditional forecasts might lead to overly confident anticipation of how that agent may influence the predicted agents \cite{InterventionalBehaviorPrediction}. To include domain knowledge such as system dynamics into the learning process, it is also common practice \cite{MultiPathPlusPlus, DKM, TrajectronPP} to first forecast the future control values of all agents and then to unroll a dynamics model to produce the future states.

\textbf{Energy-based Models.} 
The work of Xie et al. \cite{XieICML2016} proposes parametrizing an EBM with a neural network, and learning is performed with type (i) MLE. \cite{Du2020} uses an EBM for model-based single-agent planning. Our work is also related to the cooperative training paradigm \cite{xieICLR, XieTAMI2020, XiAAI2021, XieTAMI2021}, in which a  (fast-thinking) latent variable model and a (slow-thinking) EBM are trained together. These works parameterize the latent variable model used initializing the EBM, with a generator \cite{XieTAMI2020}, a variational auto-encoder \cite{XiAAI2021}, or a normalizing flow \cite{xieICLR}. 
In contrast, our work does not separate the training of the initialization and the EBM with different networks.

\textbf{Differentiable Optimization for Motion Planning} Differentiable optimization has also been applied in  motion planning for SDVs. \citet{WeiTRO} and \citet{diehl2022differentiable} impose safety-constraints using differentiable control barrier functions or gradient-based optimization techniques in static environments. \citet{Diffstack} and \citet{huang2023differentiable} couple a differentiable single-agent motion planning module with learning-based motion forecasting modules. In contrast, our work performs multi-agent joint optimizations in parallel, derived from a game-theoretic potential game formulation. Game-theoretic formulations can overcome overly conservative behavior when used for closed-loop control \citet{liu2023learning}.

\section{Theorems}
\label{app:theorem}
This section provides the full theorem of \cite{PotentialILQR}:
\begin{theorem}
	For a differential game $\Gamma_{\mathbf{x}_0}^T:=\left(T,\left\{\mathbf{u}_i\right\}_{i=1}^N,\left\{C_i\right\}_{i=1}^N,f\right)$, if for each agent $i$, the running and terminal costs have the following structure
$L_i(\mathbf{x}(t), \mathbf{u}(t), t)=p(\mathbf{x}(t), \mathbf{u}(t), t)+c_i\left(\mathbf{x}_{-i}(t), \mathbf{u}_{-i}(t), t\right)$ and
$$
S_i(\mathbf{x}(T))=\bar{s}(\mathbf{x}(T))+s_i\left(\mathbf{x}_{-i}(T)\right),
$$
then, the open-loop control input $\mathbf{u}^*=\left(\mathbf{u}_1^*, \cdots, \mathbf{u}_N^*\right)$ that minimizes the following 
$$
\begin{aligned}
	\min _{u(\cdot)} & \int_0^T p(\mathbf{x}(t), \mathbf{u}(t), t) d t+\bar{s}(\mathbf{x}(T)) \\
	\text { s.t. } & \dot{x}_i(t)=f_i\left(\mathbf{x}_i(t), \mathbf{u}_i(t), t\right),
\end{aligned}
$$
is an OLNE of the differential game $\Gamma_{\mathbf{x}_0}^T$, i.e., $\Gamma_{x_0}^T$ is a potential differential game.
\end{theorem}
Proof: See \cite{PotentialILQR}, with original proof provided by \citet{PotentialDifferentialGames}.

Here besides the potential functions $p$ and $\bar{s}$, $s_i$ and $c_i$ are terms that are required to not depend on the state or control of agent $i$.

\section{Datasets}
\label{app:datasets}
\subsection{RPI}  
\begin{figure}
	\vspace{-1cm}
	\label{fig::ilq_scenario}
	\includegraphics[width=\columnwidth]{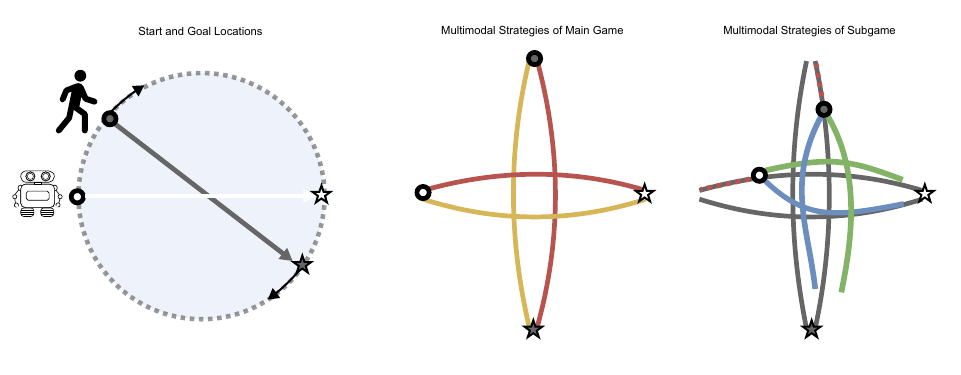}
	\vspace{-1cm}
	\caption{Dataset construction for RPI dataset. Left: First initial and goal states and game parameters are sampled. Middle: A \textit{main game} is solved multiple times based on the sampled game configuration with subsequent result clustering. That leads to multimodal strategies (red and yellow). The agent moves according to the multimodal strategies of the main game. After a time step $\Delta t$, a \textit{sub game} is solved. The results are multimodal strategies (blue and green) of the subgame. The histories (dotted red) and multi-modal strategies of the sub game build a demonstration for training and evaluation.}
\end{figure}

The RPI dataset is a synthetic dataset of simulated mobile robot pedestrian interactions. Multi-modal demonstrations are generated by approximately solving a two-player differential game ($N=2$) with the iterative linear-quadratic game implementation of \citet{ILQG2020} based on different start and goal configurations. Fig. 7 \ref{fig::ilq_scenario} provides an illustration for the dataset construction. The robot's initial positions (white circle) and goal locations (white star) are the same in all solved games. In contrast, the initial state (dark grey circle) and goal location (dark grey stars) of the pedestrian move on a circle, as illustrated on the left graphic in Fig. 7 \ref{fig::ilq_scenario}. As solving the game once leads to a unimodal local strategy, this work follows the implementation of \citet{Peters2020}. It solves the PDG for a given initial configuration multiple times based on different initializations. Afterward the resulting strategies are clustered. The clustered strategies represent multi-modal strategies of the \textit{main game}, and they are visualized in red and yellow in Figure \ref{fig::ilq_scenario}. The agents are tasked to reach a goal location given an initial start state while avoiding collisions and minimizing control efforts. The agents then execute the open-loop controls of the main game's initial strategies. After every time interval $\Delta t=0.1$, the procedure of game-solving and clustering the results are repeated as long as the agents pass each other. The resulting strategies of the so-called \textit{subgames} are visualized in green and blue on the right of Fig. 7 \ref{fig::ilq_scenario}. Based on the history (dotted red line) and the strategies of the subgame (blue and green), we then build a multi-modal demonstration for the dataset. Note that the main game and the corresponding subgames use the same cost function parametrizations, but the agents' preferences for collision avoidance differ between main games.

The resulting dataset is based on 20 main games and their corresponding subgame solutions. Here we draw collision cost parameters from a uniform distribution to enhance demonstration diversity. The resulting dataset contains 60338 samples, whereas we use 47822 ($\sim$80\%) for training, 6228 for validation ($\sim$10\%), and 6228 ($\sim$10\%) for testing. The test set is constructed based on an unseen main game configuration. The goal is to predict $M=2$ joint futures of $T=\SI{4}{\second}$ based on a history of $H=\SI{1.8}{\second}$ with a time interval of $\Delta t = 0.1$.

\begin{figure}
	\includegraphics[width=\textwidth]{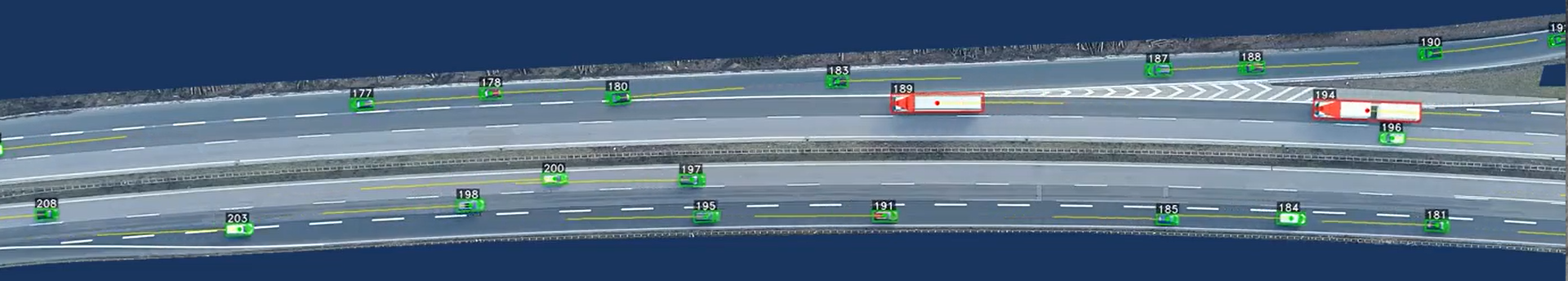}
	\label{fig::exid_scenario}
	\caption{An exemplary highly-interactive scenario from the exiD dataset.}
	\vspace{-0.5cm}
\end{figure}

\subsection{exiD} The exiD \cite{exiD2022} dataset contains \SI{19}{\hour} of real-world highly interactive highway data. Interactions between different type of vehicle classes (vehicles) are rich because the data was recorded by drones flying over seven locations of German highway entries and exits. Highway entries and exits, designed with acceleration and deceleration lanes and high-speed limits, promote interactive lane changes due to high relative speeds between on-ramping and remaining road users. In addition, the most common cloverleaf interchange in Germany requires simultaneous observation of several other road users and gaps between them for safe entry or exit in a short time frame \cite{exiD2022}. 

To further increase the interactivity, this work extracts scenarios with $N=4$ agents in which at least one agent performs a lane change. We choose $N=4$ as this resulted in the highest number of samples assuming a fixed number of agents. The recordings are then sampled with a frequency of $\Delta t = $ \SI{0.2}{\second}. The different networks (see Section \ref{Experiments}) are tasked to predict $M=5$ joint futures of length $T=\SI{4}{\second}$ based on a history of $H=\SI{1.8}{\second}$. The resulting dataset contains 290735 samples, whereas we use 206592 ($\sim$72\%) for training, 48745 for validation ($\sim$16\%), and 35398 ($\sim$12\%) for testing. To investigate the generalization capabilities of the different models, the test set contains \textit{unseen scenarios from a different map} (map 0) than the training and validation scenarios.
An exemplary scenario is visualized in Fig. 8 \ref{fig::exid_scenario}.

\section{Implementation Details}
\label{app:implementation_details}
This section provides additional information for the used observation encoding backbones and the game parameter decoders.
\subsection{Network Architectures}
\label{app:network_structure}
\textbf{Lane Encoder.}
In all experiments, the lane encoder $\phi^{\textrm{lane}}$ uses a PointNet \cite{PointNet} like architecture as \citet{VN_2020_CVPR} with three layers and a width of 64. The polylines are constructed based on vectors that contain a 2-D start and 2-D goal position in a fixed-global coordinate system. Agent polylines also include time step information and are encoded with different encoders depending on the used backbone.

\textbf{Agent History Encoder.}
The V-LSTM (Vector-LSTM) \cite{Ettinger_2021_ICCV} and VIBES (Vectorized Interaction-based Scene Prediction) backbones use an LSTM \cite{LSTM} for agent history encoding with depth three and width 64. 
Our modified HiVT-M (Hierarchical Vector Transformer Modified) \cite{HiVT} implementation uses a transformer \cite{Attention} for the encoding of each agent individually. Note that this contrasts with the original implementation, where the encoding transformer already models local agent-to-agent and agent-to-lane interactions. We account for that in a modified global interaction graph as listed below. The transformer has a depth of three and a width of 64.

\textbf{Global Interaction.}
The V-LSTM backbones update the polyline features in the global interaction graph with a single layer of attention \cite{Attention} as described by \citet{VN_2020_CVPR}. The HiVT-M and VIBES models use a two-stage attention mechanism. First, one layer of attention between the map and agent polyline features, and afterwards a layer of attention between all updated agents features are applied.
The global interaction graph has a width of 128.

\textbf{Game Parameter and Initial Strategy Decoder.}
The agent weight, goal, and initial strategy decoders are implemented by a 3-layer MLP with a width of 64.

\textbf{Goal Decoder.} 
The goal decoder follows \citet{TNT}. It takes as input the concatenation of an agent feature $\mathbf{z}_i$ and $G=60$ possible goal points, denoted by $\mathbf{z}^{\textrm{goal}}$. The goal points are extracted from the centerlines of the current and neighboring lanes. If there exists no neighboring lane, we take the lane boundaries. The decoder $\phi^{\textrm{goal}}$ then predicts the logits of a categorical distribution per agent   $\mathbf{l}^{\textrm{goal}}_{i}=\phi^{\textrm{goal}}(\mathbf{z}^{\textrm{goal}})$. During training and evaluation; the method samples the $M$ most-likely goals to receive goals $\mathbf{G}_i$ for all modes of a agent $i$. Probabilities for the goals per agents are computed by $\mathbf{PR}_i^{\textrm{goal}} = \text{softmax}(\mathbf{l}_i^\textrm{goal})$.
The prediction of goals is made in parallel for all agents.

\textbf{Scene Probability Decoder.}
The scene probability decoder also uses a 2-layer MLP with width $16 \times M$ and predicts logits $\mathbf{l}^\textrm{prob}$ for the $M$ scene mode. The scene probabilities are derived by applying the softmax operations $\mathbf{PR} = \text{softmax}(\mathbf{l}^\textrm{prob})$.  

The goal, agent weight and scene probability decoder use batch normalization. The interaction weight decoder, initial strategy decoder, and transformer agent encoder use layer normalization.

\subsection{Dynamics} 
\label{app:dynamics}
The discrete-time dynamically-extended unicycle dynamics \citet[Chapter~13]{lavalle} are given by:
\begin{center}
	\centering
	\begin{minipage}{.45\textwidth}
		\begin{equation}
			\begin{aligned}
				x_{k+1} & =x_k+v_k \cos \left(\theta_t\right) \Delta t \\
				y_{k+1} & =y_k+v_k \sin \left(\theta_t\right) \Delta t \\
				v_{k+1} & =v_k+a_k \Delta t \\
				\theta_{k+1} & =\theta_k+\omega_k \Delta t
			\end{aligned}\label{eq:unicycle}
		\end{equation}
	\end{minipage}\hfill

\end{center} $x_k$ and $y_k$ denote a 2-D position and $\theta_k$ the heading in fixed global coordinate system. $v_k$ is the velocity, $a_k$ the acceleration, $\omega_k$ the turnrate, $\delta_k$ the steering angle and $\Delta t$ a time interval. Hence, $n_{x}=4\times N$ and $n_{u}=2\times N$.

\subsection{Energy Features and Optimization}
\textbf{Energy Features.} The energy function in the RPI experiment uses the following agent-dependent features: $c(\cdot)=\left[c_{\textrm{goal}}, c_{\textrm{vel}}, c_{\textrm{acc}},  c_{\textrm{velb}}, c_{\textrm{accb}}, c_{\textrm{turnr}}, c_{\textrm{accb}}, c_{\textrm{turnrb}}\right]$. In the RPI experiments, the goal is given and not predicted.
The agent-dependent energy features in the exiD experiments are given by $c(\cdot)=\left[c_{\textrm{goal}}, c_{\textrm{lane}}, c_{\textrm{velref}}, c_{\textrm{vel}}, c_{\textrm{acc}}, c_{\textrm{jerk}}, c_{\textrm{steer}}, c_{\textrm{turnr}}, c_{\textrm{turnacc}} \right]$. $c_{\textrm{goal}}$ is a terminal cost penalizing the position difference of the last state to the predicted goal. $c_{\textrm{lane}}$ minimizes the distance of the state trajectory to the reference lane to which the predicted goal point belongs. Note that different goal points can be predicted for the modes. Hence different lanes can be selected to better model multi-modality. $c_{\textrm{velref}}$ is the difference between the predicted and map-specific velocity limits. The other terms are running cost, evaluated for all timesteps and penalize high velocities  ($c_{\textrm{vel}}$), accelerations ($c_{\textrm{acc}}$), jerks ($c_{\textrm{jerk}}$), as well as turn rates ($c_{\textrm{turnr}}$) and turn accelerations ($c_{\textrm{turnacc}}$). An index \texttt{b} marks a soft constraint implemented as a quadratic penalty, active when the constraint is violated. Hence a inequality constraint $g(z)\leq 0$ with optimization variable $z$ is implemented by a feature $\max(0, g(z))$. The interaction feature $d(\cdot)$ is also implemented as such a quadratic penalty. We evaluate the collision avoidance features at every discrete time step in the RPI experiments. In both experiments, agent geometries are approximated by circles of radius $r_i$, which is accurate for the mobile robot and pedestrian but an over-approximation for vehicles and especially trucks in the highway exiD environment, where we use $r_i=L/2$. $L$ is the length of a vehicle. Future work could also use more accurate vehicle approximations (e.g., multiple circles \cite{zielger2010}) to further evaluate collision avoidance at every time step to increase the predictive performance at a higher runtime and memory cost. In the RPI experiments, we set $r_i=\SI{0.25}{\metre}$.

\textbf{Optimization.} As the approach already predicts accurate initial strategies  $\mathbf{U}^{\textrm{init}}$, our experiments only required a few optimization steps. Concretely, the results of Tab. \ref{tab:RPI} and \ref{tab:exidmaineval} in the main paper use $s=2$ optimization steps, rendering our approach real-time capable (see Fig. \ref{fig::runtime}). Note while the approach also works, with a  higher number of optimization steps (see Fig. \ref{fig::ablationsteps}), our experiments showed that fewer optimization steps lead to similiar results, with decreased runtime and memory requirements due to the predicted initialization. Both experiments use a stepsize of $\alpha=0.3$. The experiments use a damping factor of $dp=10$ in the Levenberg-Marquardt solver \cite{Theseus}.

\subsection{Training Details}

\textbf{Loss Functions.}
The imitation loss in our experiments is the minSADE \cite{ImplicitLatentVariabModel, weng2023joint} given by: 
\begin{equation}
	\mathcal{L}^{\textrm{imit}}= \min _{m=1}^M \frac{1}{N}\sum_{i=1}^N\left\|\mathbf{x}_i^{m}-\mathbf{x}_\textrm{GT}\right\|^2
\end{equation}
It first calculates the average over all distances between agent trajectories $\mathbf{x}_i^{m}$ from agent $i$ and mode $m$ and the ground truth $\mathbf{x}_\textrm{GT}$. Then the minimum operator is applied to afterwards backpropagate the difference of the joint scene, which is closest to the ground truth.
The second loss term $\mathcal{L}^{\textrm{goal}}$ computes the cross entropy $\mathrm{CE}$ for the goal locations averaged over all agents
\begin{equation}
	\mathcal{L}^{\textrm{goal}}= \frac{1}{N}\sum_{i=1}^N\mathrm{CE}\left(\mathbf{PR}^{\textrm{goal}}_i,\mathbf{g}_i^*\right),
\end{equation} whereas $\mathbf{g}_i^*$ is the goal location of the set of $G$ possible goals closest to the ground truth goal location.
Lastly,  $\mathcal{L}^{\textrm{prob}}$ computes the cross entropy for the joint futures
\begin{equation}
	\mathcal{L}^{\textrm{prob}}= \mathrm{CE}\left(\mathbf{PR},\mathbf{x}^*\right),
\end{equation} whereas $\mathbf{x}^*$ is the predicted joint 2-D position trajectory, which has the smallest distance (measured by minSADE) to the future ground truth joint 2-D position trajectory.
We empirically set $\lambda_1=1, \lambda_2=0.1, \lambda_3=0.1$ in Eq. (\ref{equ::loss}).

\label{app::losses}
All approaches are trained with batch size 32, using the Adam optimizer \cite{Adam}. Our models in both experimental environments use a learning rate of 0.00005 across all backbones. Note that the evaluation favors the baselines, as we performed grid searches for their learning rates, whereas our approach uses the same learning rate across all backbones. 

Training and evaluation was performed using an AMD Ryzen 9 5900X and a Nvidia RTX 3090. 

\section{Additional Experiments}

\subsection{Qualitative Results}
\label{app:timeseries}
This section provides extended qualitative results. 

\begin{figure}
	\label{fig::rpi_comp_qualitative}
	\centering
	\includegraphics[width=1.0\textwidth]{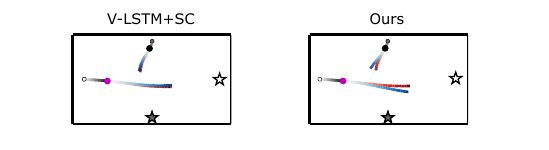}
	\vspace{-0.7cm}
	\caption{Qualitative comparison of the multi-modal ($M=2$) joint predictions in the RPI environment. The start and end point of the pink agent are located on a circles with a radius of 3m . The start and endpoint of the black agent are visualized with a grey circle and star. The different modes are visualized in red and blue color. The start and endpoints are located on a circles with a radius of three meter.}
\end{figure}

\textbf{RPI.} Fig. 9 \ref{fig::rpi_comp_qualitative} visualizes an exemplary qualitative result of the RPI experiments. Both modes collapsed when using the V-LSTM+SC baseline (explicit strategy). In contrast, this work's implicit approach better models the multi-modality present in the demonstration. Since the dataset contains solutions of games solved with different collision-weight configurations, it can be seen that our proposed method accurately differentiates between different weightings of collisions, as can be seen in the trajectories. This finding aligns with these of \citet{Florence}, which discovered that implicit models could better represent the multi-modality of demonstrations.

\begin{figure}
	\label{fig::modes_comp_exid_1}
	\centering
	\includegraphics{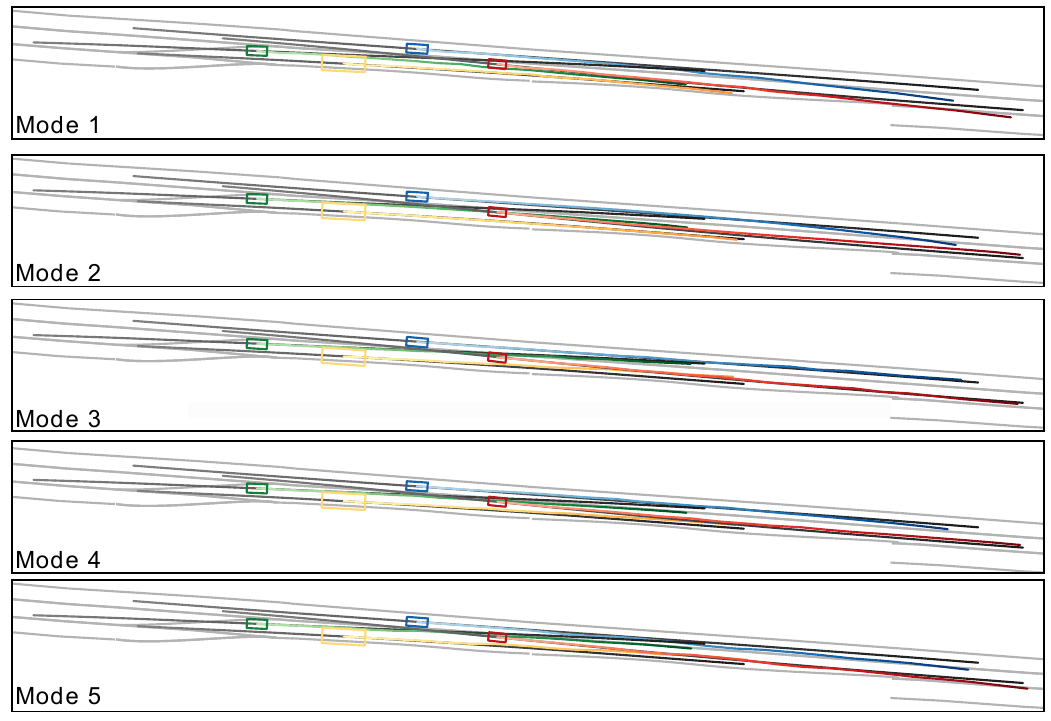}
	\vspace{-0.5cm}
	\caption{Multi-modal predictions in an interactive scenario, where the green and yellow perform on-ramp merges. The agent trajectories are visualized in different colors, whereas the color changes with an increasing number of predicted steps. The ground truth (history and future) is shown with colors from dark grey to black and the map in light grey.}
\end{figure}

\begin{figure}
	\label{fig::modes_comp_exid_2}
	\centering
	\includegraphics{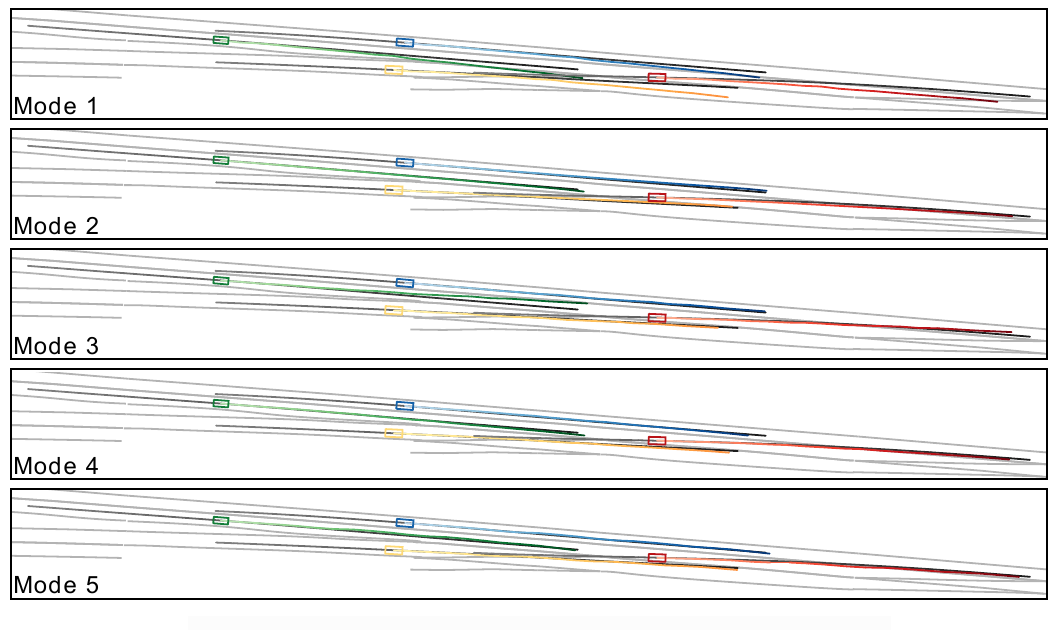}
	\vspace{-0.5cm}
	\caption{Multi-modal predictions in a interactive scenario, where the yellow and red agents perform lane changes. The agent trajectories are visualized in different colors, whereas the color changes with an increasing number of predicted steps. The ground truth (history and future) is shown with colors from dark grey to black and the map in light grey.}
\end{figure}

\textbf{exiD.} Fig. 10 \ref{fig::modes_comp_exid_1} visualizes multi-modal predictions in a highly interactive scenario, where one car (green) and one truck (yellow) merge onto the highway. The green car performs a double-lane change. Note how our model in mode three accurately predicts the future scene evolution and also outputs reasonable alternative futures. For example, in mode one, the green car performs a single lane change, whereas the blue and red cars are also predicted to change lanes.
Another multi-modal prediction is visualized in Fig. 11 \ref{fig::modes_comp_exid_2}. Observe again how the ground truth is accurately predicted in this interactive scenario (mode 5), whereas, for example, also other plausible futures are generated. For instance, the yellow vehicle stays longer on the acceleration lane in mode one, whereas in mode three, the green vehicle performs a lane change.

\end{document}